\documentclass{article}

\PassOptionsToPackage{numbers}{natbib} %,compress?
\usepackage[preprint]{neurips_2021}

\usepackage[utf8]{inputenc} % allow utf-8 input
\usepackage[T1]{fontenc}    % use 8-bit T1 fonts
\usepackage{hyperref}       % hyperlinks
\usepackage{url}            % simple URL typesetting
\usepackage{booktabs}       % professional-quality tables
\usepackage{amsfonts}       % blackboard math symbols
\usepackage{nicefrac}       % compact symbols for 1/2, etc.
\usepackage{microtype}      % microtypography
\usepackage{xcolor}         % colors

\usepackage{amsmath} %align envs
\usepackage{bm}
\usepackage{subfig} %for subfloat
\usepackage{graphicx}

\usepackage{floatrow}
\newfloatcommand{capbtabbox}{table}[][\FBwidth]

\usepackage{wrapfig}

\title{Bayesian Classifier Fusion with an Explicit Model of Correlation}

\author{%
  Susanne Trick \\
  Centre for Cognitive Science \& Institute of Psychology\\
  TU Darmstadt\\
  Darmstadt, Germany \\
  \texttt{susanne.trick@cogsci.tu-darmstadt.de} \\
  \And
  Constantin A. Rothkopf \\
  Centre for Cognitive Science \& Institute of Psychology \\
  TU Darmstadt \\
  Darmstadt, Germany \\
  Frankfurt Institute for Advanced Studies \\
  Goethe University Frankfurt \\
  Frankfurt, Germany \\
  \texttt{constantin.rothkopf@cogsci.tu-darmstadt.de} \\
}

\begin{document}

\maketitle

\begin{abstract}
Combining the outputs of multiple classifiers or experts into a single probabilistic classification is a fundamental task in machine learning with broad applications from classifier fusion to expert opinion pooling. Here we present a hierarchical Bayesian model of probabilistic classifier fusion based on a new correlated Dirichlet distribution. 
This distribution explicitly models positive correlations between marginally Dirichlet-distributed random vectors thereby allowing explicit modeling of correlations between base classifiers or experts.
The proposed model naturally accommodates the classic Independent Opinion Pool and other independent fusion algorithms as special cases. It is evaluated by uncertainty reduction and correctness of fusion on synthetic and real-world data sets. We show that a change in performance of the fused classifier due to uncertainty reduction can be Bayes optimal even for highly correlated base classifiers.
\end{abstract}

%%%%%%%%%%%%%%%%%%%%%%%%%%%%%%%%%%%%%%%%%%%%%%%%%%%%%%%%%%%%%%%%%%%%%%%%%%%%%%%%%%%%%
%%%%%%%%%%%%%%%%%%%%%%%%%%%%%%%%%%%%%%%%%%%%%%%%%%%%%%%%%%%%%%%%%%%%%%%%%%%%%%%%%%%%%
\section{Introduction} \label{sec:introduction}
Classification is one of the fundamental tasks in machine learning with broad applicability in many domains.
The most successful classification methods, e.g. in machine learning competitions, have proven to be classifier ensembles, which combine different classifiers to improve classification performance \citep{kittler1998, dietterich2000, mohandes2018, pirs2019}. 
Apart from the selection and training of individual classifiers, the fusion method used for classifier combination is of particular importance for the success of an ensemble, as individual classifiers can be biased or highly variable.
Such fusion methods can equivalently be applied for fusing human experts' opinions.
However, for convenience, most common fusion methods assume independent classifiers \citep{schubert2004, mohandes2018}, although in practice, classifiers trained on the same target as well as human experts are highly correlated \citep{jacobs1995}. 

Different strategies for coping with correlated classifiers have been proposed, such as selecting only those classifiers with the lowest correlation \citep{petrakos2000, prabhakar2002, goebel2004, faria2013, singh2018}, explicitly decorrelating the classifiers before fusion \citep{ulas2012}, or weighting them according to their correlation \citep{srinivas2009, terrades2009, lacoste2014, safont2019}.
While there are several non-Bayesian models of improved fusion of correlated classifiers \citep{drakopoulos1988, kam1991, baertlein2001, veeramachaneni2008, sundaresan2011, ma2013}, \citet{kim2012} introduced a Bayesian model for fusing dependent discrete classifier outputs, albeit not probabilistic outputs, thereby disregarding valuable information about the uncertainty of decisions.
\citet{pirs2019} extend the work of \citet{kim2012} by allowing probabilistic classifier outputs. But, their focus is on outperforming related fusion algorithms 
using an approximate model of dependent classifiers
rather than developing a theoretically justified normative model of how correlated classifier fusion should work. 
In particular, \citet{pirs2019} conclude that
a fusion method should not outperform the base classifiers if these are highly correlated.
However, while it is known that there should be no fusion gain for a correlation of $r=1$ between classifiers \citep{drakopoulos1988, tumer1996, kuncheva2000, petrakos2000, baertlein2001, zhou2012}, this has not been shown for probabilistic classifiers.
Here, we clarify how the correlation between classifiers affects uncertainty reduction through fusion in general, which is well known in the case of fusing independent probabilistic classifier outputs \citep{andriamahefa2017}.

Therefore, in order to show how correlated probabilistic classifier outputs should be fused Bayes optimally, in this work we introduce a hierarchical fully Bayesian normative model of the fusion of correlated probabilistic classifiers. We model the classifiers to be fused with a new correlated Dirichlet distribution, which is able to model Dirichlet-distributed random vectors with positive correlation.
We show that existing fusion methods such as Independent Opinion Pool are special cases of this model.
Evaluations on simulated and real data reveal that fusion should reduce uncertainty the less, the higher the classifiers are correlated. In particular, if the classifiers' correlation is 1, there should be no uncertainty reduction through fusion. Still, since we learn a model of each base classifier, this does not necessarily mean that the fused distribution equals the base distributions. Empirical evaluations show the approach's superiority on real-world fusion problems.

%%%%%%%%%%%%%%%%%%%%%%%%%%%%%%%%%%%%%%%%%%%%%%%%%%%%%%%%%%%%%%%%%%%%%%%%%%%%%%%%%%%%%
%%%%%%%%%%%%%%%%%%%%%%%%%%%%%%%%%%%%%%%%%%%%%%%%%%%%%%%%%%%%%%%%%%%%%%%%%%%%%%%%%%%%%
\section{Related work} \label{sec:related_work}
Bayesian models of classifier fusion are known as Supra-Bayesian fusion approaches \citep{jacobs1995}. For combining expert opinions, they have already been proposed before machine learning methods emerged. Considering the opinions as data, a probability distribution is learned over them, conditional on the true outcome. From this expert model, a decision maker can compute the likelihood of observed opinions and combine it with its prior using Bayes' rule. The resulting posterior distribution over the possible outcomes is the fusion result \citep{genest1986}.
For instance, \citet{lindley1985}, \citet{french1980}, and \citet{winkler1981} modeled experts' opinions using a multivariate normal distribution, which enabled explicit modeling of their correlations, while \citet{jouini1996} used copulas to model experts' correlations.

Such Supra-Bayesian approaches have also been proposed for classifier fusion. \citet{kim2012} model independent discrete classifier outputs by learning a multinomial distribution over each row of the classifiers' confusion matrices, conditioned on the true class label. This Independent Bayesian Classifier Combination Model (IBCC) is additionally extended to a Dependent Bayesian Classifier Combination Model (DBCC), which uses Markov networks to model correlations. Inference is realized with Gibbs Sampling, and training is unsupervised.
Several authors have extended the work of \citet{kim2012}. However, most of them extend the IBCC method, which assumes independent classifiers. For example, \citet{simpson2013} infer the IBCC parameters with variational inference instead of Gibbs Sampling. \citet{hamed2018} instead presented a supervised extension of IBCC.
\citet{ueda2014} additionally introduce another latent variable into the original IBCC model that determines a classifier's effectiveness, i.e. whether it always outputs the same label for a class or varies considerably. 
Still, as in \citep{kim2012}, this line of work considers discrete classifier outputs without utilizing classifiers' uncertainties for fusion.
Thus, \citet{nazabal2016} introduced a Bayesian model for fusing probabilistic classifiers that output categorical distributions instead of only discrete class labels. The output distributions of each classifier are modeled with a Dirichlet distribution conditioned on the true class label. Parameter inference is realized with Gibbs Sampling on labeled training data. However, similar to the approaches above, the model assumes independent base classifiers and disregards potential correlations.

In contrast, \citet{pirs2019} explicitly model correlations between probabilistic classifiers. They transform the classifiers' categorical output distributions with the inverse additive logistic transform and model the resulting real-valued vectors with mixtures of multivariate normal distributions with means and covariances conditioned on the true class labels. While \citet{pirs2019} show that this model outperforms other Bayesian fusion methods on most data sets, the model does not provide a normative account of how fusion of correlated probabilistic classifiers should work Bayes optimally. In particular, they conclude
that a fused classifier cannot outperform the base classifiers if these are highly correlated and provide empirical evidence for this conclusion based on one data set. 
However, this has not been proven for probabilistic classifiers, where a special focus should be on uncertainty reduction through fusion. 
To investigate how this uncertainty reduction should be affected by correlation, we propose a normative hierarchical Bayesian generative model of the fusion of correlated probabilistic classifiers. 
The model's structure resembles the structure presented by \citet{pirs2019} 
up to a newly introduced conjugate prior of the categorical distribution, a correlated Dirichlet distribution for jointly modeling the classifier outputs.
In contrast to \citet{pirs2019}, 
we do not require any transformation of the classifier outputs or mixture distributions and show that the fused classifier can outperform the base classifiers, even for highly correlated base classifiers.

%%%%%%%%%%%%%%%%%%%%%%%%%%%%%%%%%%%%%%%%%%%%%%%%%%%%%%%%%%%%%%%%%%%%%%%%%%%%%%%%%%%%%
%%%%%%%%%%%%%%%%%%%%%%%%%%%%%%%%%%%%%%%%%%%%%%%%%%%%%%%%%%%%%%%%%%%%%%%%%%%%%%%%%%%%%
\section{Bayesian models of classifier fusion} \label{sec:approach}
Throughout this work, we assume $K$ base classifiers $C_k, k=1,...,K$ to be given and fixed. For a given example $i$, each base classifier $C_k$ receives observation $o_i^k$ with corresponding true class label $t_i= 1,...,J$. Based on observation $o_i^k$, each classifier $C_k$ outputs the respective probability distribution $P(t_i|o_i^k)$, which is a $J$-dimensional categorical distribution.
The goal of the present work is to fuse these given classifier outputs $P(t_i|o_i^k)$ in order to obtain $P(t_i|o_i^1, ..., o_i^K)$.
Accordingly, in the following we investigate Bayes optimal fusion methods with successively more general assumptions.
In Section \ref{sec:iop} we start with assuming independent classifiers whose behavior is not known. In Section \ref{sec:independent_fusion_model} we proceed by modeling each individual classifier's behavior while still assuming independence. The resulting Independent Fusion Model is finally extended to the Correlated Fusion Model in Section \ref{sec:correlated_fusion_model}, which explicitly models classifiers' correlations.

%%%%%%%%%%%%%%%%%%%%%%%%%%%%%%%%%%%%%%%%%%%%%%%%%%%%%%%%%%%%%%%%%%%%%%%%%%%%%%%%%%%%%
\subsection{Independent Opinion Pool} \label{sec:iop}
If we assume that the outputs of all base classifiers are conditionally independent given $t_i$ with an uninformed prior, by applying Bayes' rule we can transform the sought $P(t_i|o_i^1, ..., o_i^K)$ to:
\begin{align} \label{eq:iop_bayes_rule}
P(t_i|o_i^1, ..., o_i^K)
&\propto \prod_{k=1}^{K}{P(t_i|o_i^k)},
\end{align}
which needs to be renormalized to sum to 1.
This fusion rule, which is known as Independent Opinion Pool (IOP) \citep{berger1985}, is therefore Bayes optimal given the stated assumptions.
Also, it leads to intuitive results regarding uncertainty. Non-conflicting base distributions reinforce each other in a way that the fused categorical distribution's uncertainty is reduced \citep{andriamahefa2017}, and the more uncertain a base distribution, the less it affects the resulting fused distribution \citep{hayman2002}.

%%%%%%%%%%%%%%%%%%%%%%%%%%%%%%%%%%%%%%%%%%%%%%%%%%%%%%%%%%%%%%%%%%%%%%%%%%%%%%%%%%%%%
\subsection{Independent Fusion Model} \label{sec:independent_fusion_model}
Although IOP is Bayes optimal given conditionally independent base classifiers and an uninformed prior, it is an ad-hoc method. Thus, only information given by the current output distributions can be exploited for fusion. The individual classifiers' properties, their bias, variance, and uncertainty, cannot be considered. Therefore, the Independent Fusion Model (IFM) additionally models the behavior of the classifiers to be fused, while still assuming conditional independence of classifiers and an uninformed prior over classes.
Since modeling each classifier's behavior requires considering their categorical output distributions as data, here we assume them as given and fixed and define them as $\bm{x_i^k} = P(t_i|o_i^k)$ for base classifier $C_k$ and example $i$.

By observing multiple training examples of classifier outputs $\bm{x_i^k}$, a probability distribution over them conditional on the true class label $t_i$ can be learned, $P(\bm{x_i^k}|t_i)$. We set this distribution to be a Dirichlet distribution. Thus, if $t_i$ can take $J$ different values, each base classifier's outputs are modeled by $J$ Dirichlet distributions, $P(\bm{x_i^k}|t_i=1), ..., P(\bm{x_i^k}|t_i=J)$.
The graphical model of the proposed IFM is shown in Figure \ref{fig:fusion_models}(a). The true label $t_i$ of example $i$ is modeled with a categorical distribution with parameter $\bm{p}$. If sufficient knowledge about the data is available, the prior $\bm{p}$ over true labels $t_i$ can be chosen accordingly. For the subsequent experiments we chose an uninformed prior with $\bm{p}=(\frac{1}{J}, ..., \frac{1}{J})$.
$\bm{\alpha}$ holds the parameters of the Dirichlet distributions that model the classifiers' outputs. $\bm{\alpha_j^k}$ with ${\alpha_j^k}_l > 0$ for $l=1,...,J$ thereby contains the parameters of the Dirichlet distribution over the outputs of classifier $C_k$ if $t_i=j$. Hence, the output $\bm{x_i^k}$ of classifier $C_k$ for example $i$ with true label $t_i=j$ is Dirichlet-distributed with parameter vector $\bm{\alpha_j^k}$.

A similar model was proposed by \citet{nazabal2016}. However, their model uses more parameters since they chose the parameters of Dirichlet distributions to be a product of two parameters.

%%%%%%%%%%%%%%%%%%%%%%%%%%%%%%%%%%%%%%%%%%%%%%%%%%%%%%%%%%%%%%%%%%%%%%%%%%%%%%%%%%%%%
\subsubsection{Parameter inference} \label{sec:ind_inference}
For learning the classifier model parameters $\bm{\alpha}$, the posterior distribution over $\bm{\alpha}$ conditioned on observed classifier outputs $\bm{x}$ and the corresponding true labels $\bm{t}$, $P(\bm{\alpha}|\bm{x}, \bm{t})$, needs to be inferred.
The training data $\bm{x}$ consist of $I$ examples composed of $K$ categorical output distributions $\bm{x_i^k}$, and $\bm{t}$ holds $I$ true labels $t_i$ respectively.
Inference is performed with Gibbs Sampling. As an uninformed prior for all elements of $\bm{\alpha_j^k}$ we chose a vague gamma prior with shape and scale set to $10^{-3}$. Of course, one could choose any other prior given additional domain knowledge about the data. In the following, 
we take the expectations of inferred posterior distributions as point estimates for $\bm{\alpha_j^k}$.

%%%%%%%%%%%%%%%%%%%%%%%%%%%%%%%%%%%%%%%%%%%%%%%%%%%%%%%%%%%%%%%%%%%%%%%%%%%%%%%%%%%%%
\subsubsection{Normative fusion behavior} \label{sec:ind_fusion_behavior}
For fusion, the posterior distribution over $t_i$ given all $K$ classifier outputs $\bm{x_i^k}$ and the learned model parameters $\bm{\alpha}$, $P(t_i|\bm{x_i^1}, ..., \bm{x_i^K}, \bm{\alpha})$, needs to be inferred. Since the IFM is a generative model for independent categorical classifier outputs, performing fusion in this way is Bayes optimal given the model assumptions.
The posterior fused distribution can be derived analytically:
\begin{align}
p(t_i=j|\bm{x_i}, \bm{\alpha_j})
&\propto
\prod_{k=1}^{K}{\frac{1}{\text{B}(\bm{\alpha_j^k})}\prod_{l=1}^{J}{({x_i^k}_l)^{{\alpha_j^k}_l-1}}}. \label{eq:general_posterior}
\end{align}
This unnormalized posterior probability can now be computed for all $t_i = j$ for $j=1,...,J$, and normalizing these values to make them sum to 1 gives the posterior fused categorical distribution.

As (\ref{eq:general_posterior}) shows, using the IFM, we do not multiply the categorical output distributions of the base classifiers, such as for IOP, but their probabilities conditioned on the modeling Dirichlet distributions. 
Thus, fusion can take into account potential learned biases. Moreover, also the variances and uncertainties of the base classifiers can be considered for fusion.

This can be demonstrated with the following example. If a classifier $C_1$ is modeled by three Dirichlet distributions with parameters $\bm{\alpha_1^1} = (a+n, a, a)$ for $t_i=1$, $\bm{\alpha_2^1} = 
(a, a+n, a)$ for $t_i=2$, $\bm{\alpha_3^1} = (a, a, a+n)$ for $t_i=3$, and a classifier $C_2$ is modeled equivalently with $\bm{\alpha_1^2} = (b+m, b, b)$, $\bm{\alpha_2^2} = (b, b+m, b)$, $\bm{\alpha_3^2} = (b, b, b+m)$, with $a,b,n,m > 0$, we can simplify (\ref{eq:general_posterior}) to:
\begin{align} \label{eq:nmeq}
p(t_i=j|\bm{x_i}, \bm{\alpha_j}) &\propto ({x_i^1}_j)^{n} ({x_i^2}_j)^{m}
\end{align}
for $j=1,2,3$. This case, which was not considered by \citet{nazabal2016}, is of particular interest, because if we set parameters $n=m=1$, the IFM reduces to IOP.
However, increasing $n$ and $m$ results in lower uncertainty of the fused distribution if non-conflicting base distributions are fused. In addition, if $n>m$, $C_1$ has a higher impact on the fused result than $C_2$.

How $n$ and $m$ are related to variance and uncertainty of a classifier can be quantified with two properties of the Dirichlet distribution, its precision and the entropy of its expectation, which is a categorical distribution. The precision of a Dirichlet distribution with parameter $\bm{\alpha}$, defined as $\sum_{j=1}^{J}{\alpha_j}$, is higher, the more concentrated the distribution is around the Dirichlet's expectation \citep{huang2005}. Thus, a Dirichlet distribution with a high precision models a classifier with a low variance. On the other hand, the entropy of a Dirichlet's expectation quantifies the average uncertainty of the modeled classifier.
If we keep $a$ fixed and increase $n$, the precision of the corresponding Dirichlet distribution increases. Also, it can be shown that its expectation uncertainty decreases.
Thus, the lower classifier $C_1$'s variance and uncertainty, the higher is its fusion impact and uncertainty reduction through fusion.
If we instead increase $a$ while keeping $n$ fixed, this again increases precision and reduces $C_1$'s variance, but also increases its mean uncertainty. Hence, a classifier with low variance and high uncertainty has the same fusion impact as a classifier with high variance and low uncertainty.

Note that if we set $K=1$ in (\ref{eq:general_posterior}), the IFM can also be used as a meta classifier for a single classifier $C_1$. This meta classifier classifies a given example $i$ based on $C_1$'s output distribution $\bm{x_i^1}$. Thus, we only learn a Dirichlet model of classifier $C_1$ instead of multiple classifiers. Conditioned on the learned model parameters $\bm{\alpha^1}$ and the single base classifier's output distribution $\bm{x_i^1}$, then the posterior distribution over all possible class labels, $P(t_i=j|\bm{x_i^1},\bm{\alpha_j^1})$, is computed, which is the meta classifier's result.
\begin{figure*}
	\begin{center}
		{\subfloat[IFM]
			{\includegraphics[width=0.245\linewidth]{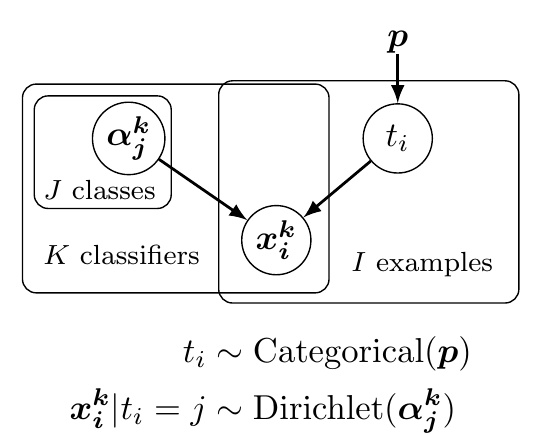}}}
		{\subfloat[CFM]
			{\includegraphics[width=0.245\linewidth]{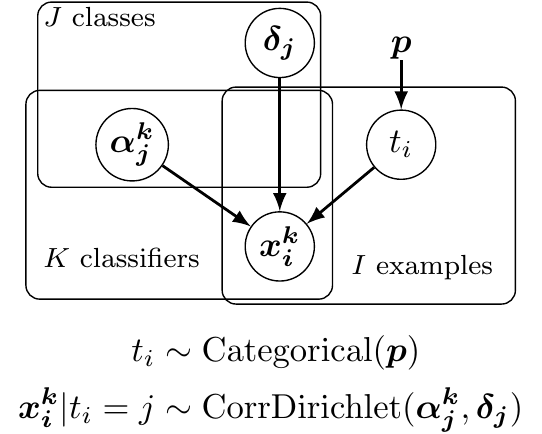}}}
		{\subfloat[CFM with latent variables of correlated Dirichlet]
			{\includegraphics[width=0.49\linewidth]{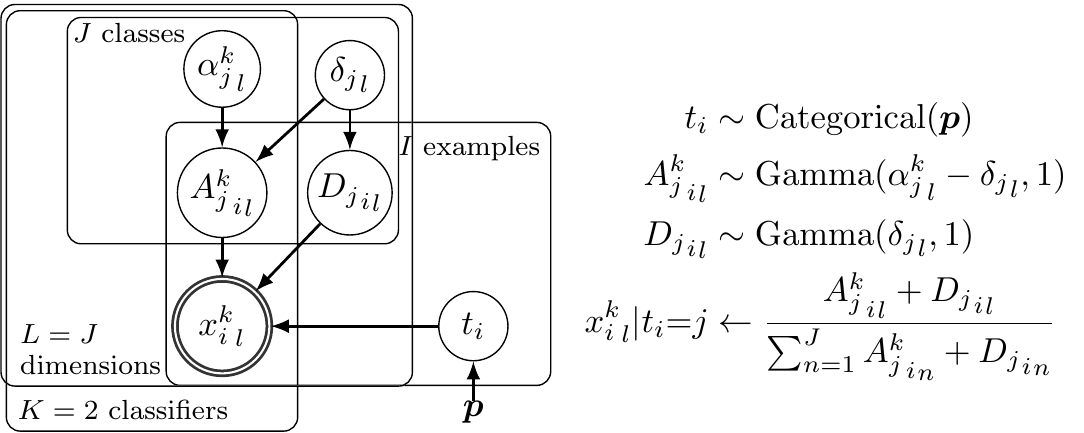}}}
		\caption{Graphical models of the IFM (a), CFM (b) and a detailed CFM for $K=2$ classifiers with all latent variables (c).}
		\label{fig:fusion_models}
	\end{center}
\end{figure*}
%

%%%%%%%%%%%%%%%%%%%%%%%%%%%%%%%%%%%%%%%%%%%%%%%%%%%%%%%%%%%%%%%%%%%%%%%%%%%%%%%%%%%%%
\subsection{Correlated Fusion Model} \label{sec:correlated_fusion_model}
The IFM introduced in Section \ref{sec:independent_fusion_model} enables optimal fusion of categorical output distributions of conditionally independent base classifiers. However, in practice most classifiers trained on the same target are highly correlated \citep{jacobs1995}. Therefore, we extend the IFM to a Correlated Fusion Model (CFM) to explicitly model the correlations between different classifiers' outputs.
As in the IFM, we also model the categorical classifier outputs $\bm{x_i^k}$ given the true label $t_i$ as a probability distribution. However, instead of modeling all classifiers independently with individual Dirichlet distributions, we model the joint distribution $P(\bm{x_i^1},..., \bm{x_i^K}|t_i)$ with a new correlated Dirichlet distribution that can express correlations between the classifiers' outputs.

%%%%%%%%%%%%%%%%%%%%%%%%%%%%%%%%%%%%%%%%%%%%%%%%%%%%%%%%%%%%%%%%%%%%%%%%%%%%%%%%%%%%%
\subsubsection{Correlated Dirichlet distribution} \label{sec:cor_dirch}
For modeling correlated classifiers' categorical output distributions with their conjugate prior, a distribution is required that can model correlations between marginally Dirichlet-distributed random variables. While previous generalizations of the Dirichlet distribution focused on more flexible \emph{correlations between individual random vector entries} $x_1,...,x_J$ of a Dirichlet variate $\bm{x}$ \citep{connor1969, wong1998, linderman2015}, here we introduce a correlated Dirichlet distribution that models \emph{correlations between two random vectors} $\bm{x^1} = (x_1^1, …, x_J^1)$ and $\bm{x^2} = (x_1^2,…,x_J^2)$ with arbitrary marginal Dirichlet distributions.

A $J$-dimensional correlated Dirichlet distribution is thereby constructed from $3J$ independent gamma variates $A_1^1, ..., A_J^1, A_1^2, ..., A_J^2, D_1, ..., D_J$ with shape parameters $\alpha_1^1 - \delta_1, ..., \alpha_J^1 - \delta_J, \alpha_1^2 - \delta_1, ..., \alpha_J^2 -\delta_J, \delta_1, ..., \delta_J$ with $\alpha_l^1, \alpha_l^2, \delta_l > 0, \alpha_l^1, \alpha_l^2 > \delta_l$, and equal scale parameter 1.
$\bm{x^1}=(x_1^1,...,x_J^1)$ and $\bm{x^2}=(x_1^2,...,x_J^2)$ with:
\begin{align} \label{eq:cor_dirch_gen}
x_l^k &= \frac{A_l^k + D_l}{\sum_{n=1}^{J}{A_n^k + D_n}},\quad l=1,...,J, k=1,2,
\end{align}
are marginally Dirichlet-distributed with Dirichlet($\bm{x^1}; \alpha_1^1, ..., \alpha_J^1$) and Dirichlet($\bm{x^2}; \alpha_1^2, ..., \alpha_J^2$). Their positive correlation, i.e. positive correlations between $x_l^1$ and $x_l^2$ for $l=1,...,J$, is generated by the shared variables $D_1,...,D_J$ with the correlation parameters $\delta_1, ..., \delta_J$. If $\delta_l$ tends to zero for $l=1,...,J$, $\bm{x^1}$ and $\bm{x^2}$ are independent and each follow a standard Dirichlet distribution. If $\bm{x^1}$ and $\bm{x^2}$ have the same marginal distributions with $\bm{\alpha^1}=\bm{\alpha^2}$, their correlation tends to 1 if $\bm{\delta}$ tends to $\bm{\alpha^1}=\bm{\alpha^2}$. Thus, if $\bm{x^1}$ and $\bm{x^2}$ have different marginal distributions, the correlation is limited below 1. While no closed-form solution for the distribution is available, sampling from it is straightforward so that it can be applied to the CFM.

Figure \ref{fig:fusion_models}(b) shows the CFM's graphical model. The only difference to the IFM in Figure \ref{fig:fusion_models}(a) is that classifier outputs $\bm{x_i^1}, ..., \bm{x_i^K}$ are jointly correlated-Dirichlet-distributed with parameters $\bm{\alpha_j^k}$ and $\bm{\delta_j}$ if $t_i=j$.
As in the IFM, $\bm{\alpha_j^k}$ with ${\alpha_j^k}_l > 0$ holds the parameter vector of the marginal Dirichlet distribution of classifier $C_k$ if $t_i = j$. The new parameter $\bm{\delta_j}$ is responsible for the pairwise correlation between the classifier outputs if $t_i=j$.
Its dimensionality is $1 \times J$ for $K=2$ and $(\binom{K}{2} +1) \times J$ for $K>2$ classifiers.
For the reduced case of $K=2$ classifiers, Figure \ref{fig:fusion_models}(c) additionally shows a more detailed graphical model of the CFM including the latent variables of the correlated Dirichlet distribution. For $K=2$, it must hold that ${\delta_j}_{l}>0$ and ${\delta_j}_{l}<{\alpha_j^k}_l$ for $l=1,...,J, k=1,...,K$.

%%%%%%%%%%%%%%%%%%%%%%%%%%%%%%%%%%%%%%%%%%%%%%%%%%%%%%%%%%%%%%%%%%%%%%%%%%%%%%%%%%%%%
\subsubsection{Parameter inference}
We learn the joint classifier model by inferring the posterior distribution over parameters $\bm{\alpha}$ and $\bm{\delta}$ given observed classifier outputs $\bm{x}$ and their true labels $\bm{t}$, $P(\bm{\alpha}, \bm{\delta}|\bm{x}, \bm{t})$, using Gibbs Sampling.
For all elements of $\bm{\alpha_j^k}$ and $\bm{\delta_j}$, we chose a vague gamma prior with shape and scale set to $10^{-3}$, which however can be set differently according to prior knowledge about the data.
To increase robustness, inference can also be split up in two steps by first inferring the marginal Dirichlet parameters $\bm{\alpha}$ as described in Section \ref{sec:ind_inference} and subsequently inferring the posterior distribution over the correlation parameters given the inferred marginal parameters, $P(\bm{\delta}|\bm{x}, \bm{t}, \bm{\alpha})$. This step-wise inference gives the same results as full inference on data generated from the CFM, but was observed to be more robust empirically on real data since it guarantees correctly inferred marginal distributions. 
As for the IFM, we use the expectation of the inferred posterior distributions as point estimates for $\bm{\alpha_j^k}$ and $\bm{\delta_j}$.

%%%%%%%%%%%%%%%%%%%%%%%%%%%%%%%%%%%%%%%%%%%%%%%%%%%%%%%%%%%%%%%%%%%%%%%%%%%%%%%%%%%%%
\subsubsection{Normative fusion behavior}
The fusion of $K$ categorical base distributions $\bm{x_i^1}, ..., \bm{x_i^K}$ is performed by inferring the posterior distribution over the true label $t_i$ conditioned on the base distributions $\bm{x_i^k}$ and the learned model parameters $\bm{\alpha}$ and $\bm{\delta}$, $P(t_i|\bm{x_i^1}, ..., \bm{x_i^K}, \bm{\alpha}, \bm{\delta})$.
Different from the IFM, here we cannot derive the fused distribution analytically because we do not have a closed-form solution for the probability density function of the correlated Dirichlet distribution. However, by assuming $\bm{\alpha}, \bm{\delta}$, and $\bm{x_i^1}, ..., \bm{x_i^K}$ to be observed, inference of latent $t_i$ can be performed with Gibbs Sampling. From a sufficient number of samples of $t_i$ we can infer the categorical distribution over $t_i$, which is the fused result. Alternatively inferring $t_i$ with variational methods in order to speed up fusion is left for future work.

Note that if we let all correlation parameters $\bm{\delta_j}$ tend to zero, the CFM reduces to the IFM, and its fusion behavior coincides with the one we derived analytically for the IFM in Section \ref{sec:ind_fusion_behavior}. Thus, bias, variance, and uncertainty of individual classifiers similarly influence the fusion when fusing with the CFM.
Additionally, in contrast to previous fusion algorithms, our model can be used to investigate how uncertainty reduction through fusion should be affected by the correlation of the fused classifiers in a normative way. We examine this in detail with the two examples in the following.

\begin{figure}
	\begin{center}
		\includegraphics[width=0.6\linewidth]{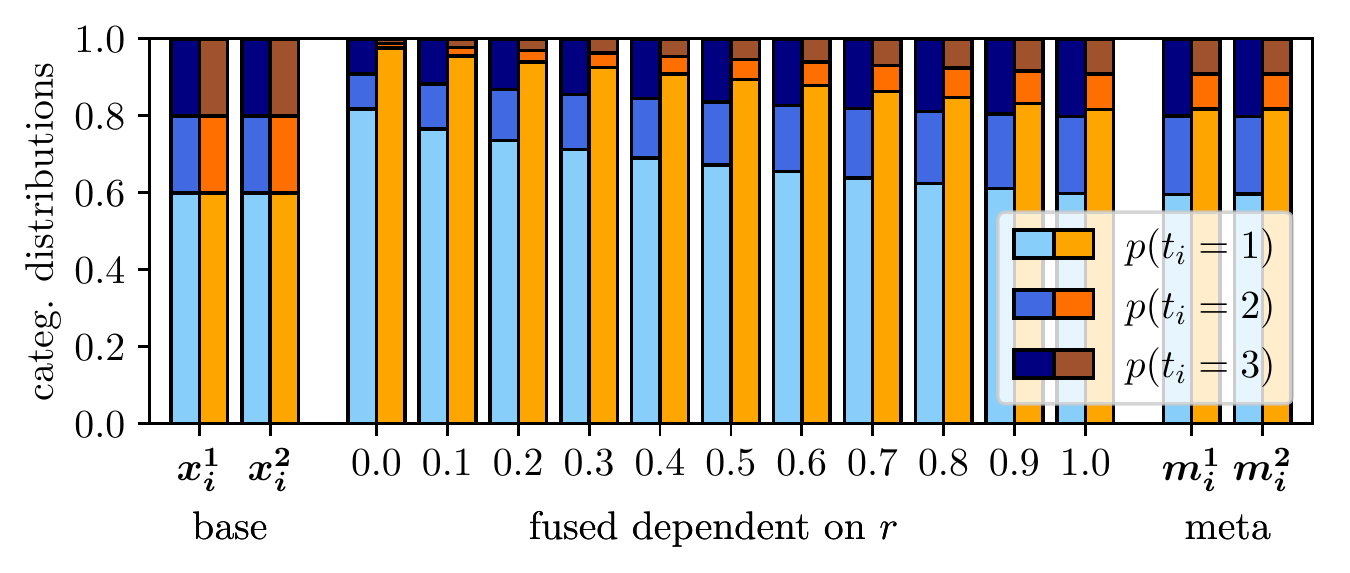}
		\caption{The base distributions $\bm{x_i^1} {=} \bm{x_i^2} {=} (0.6, 0.2 , 0.2)$ are fused using the CFM assuming IOP marginal parameters (blue bars) and marginals that imply stronger uncertainty reduction (orange bars). We progressively increase the assumed correlation between classifiers from 0.0 to 1.0 and show the corresponding fused distributions as well as the results of the meta classifiers $\bm{m_i^1}$ and $\bm{m_i^2}$.}
		\label{fig:fusion_detailed_both}
	\end{center}
\end{figure}

Specifically, we compare the fusion behavior of the CFM for systematically varied correlations between two base classifiers.
We implement inference using JAGS \citep{plummer2003}. %woanders hin?
The blue bars in Figure \ref{fig:fusion_detailed_both} show an example where the marginal parameters of the correlated Dirichlet distributions are chosen to replicate IOP fusion behavior for zero correlation ($n=m=1$ in (\ref{eq:nmeq})). The higher the correlation between the two classifiers, the smaller is the uncertainty reduction through fusion. In particular, there is no uncertainty reduction if the correlation is $r=1$. In this case, the fused distribution equals the two base distributions.
The orange bars in Figure \ref{fig:fusion_detailed_both} show the fusion results given different correlation levels for marginal parameters that imply increased uncertainty reduction compared to IOP ($n=m=2$ in (\ref{eq:nmeq})) for zero correlation because of lower classifier variance and uncertainty. As can be seen, there is also less uncertainty reduction, the higher the correlation between both classifiers. However, for $r=1$, the fused distribution is not identical to the two base distributions; its uncertainty is reduced despite the high correlation. Yet, the reason for this is not fusion but the Dirichlet models we learned for each individual classifier.
The resulting fused distribution for $r=1$ is similar to the resulting distributions we get if we use the IFM as a meta classifier individually for each base distribution (see Section \ref{sec:ind_fusion_behavior}). Hence, the fusion of two highly correlated classifiers does not additionally reduce the uncertainty. This also applies to the first example. However, in this case, due to the chosen marginal distributions, the meta classifier results are equal to the base distributions.
Both examples reveal that the uncertainty reduction through fusion should decrease progressively if the base classifiers' correlation increases. For a correlation of $r=1$, fusion should not reduce the uncertainty at all. Still, the fused distribution might be less uncertain than the base distributions since uncertainty cannot only be reduced by fusion but also as a result of modeling each individual classifier’s behavior, i.e. bias, variance and uncertainty.

%

%%%%%%%%%%%%%%%%%%%%%%%%%%%%%%%%%%%%%%%%%%%%%%%%%%%%%%%%%%%%%%%%%%%%%%%%%%%%%%%%%%%%%
%%%%%%%%%%%%%%%%%%%%%%%%%%%%%%%%%%%%%%%%%%%%%%%%%%%%%%%%%%%%%%%%%%%%%%%%%%%%%%%%%%%%%
\section{Evaluation} \label{sec:evaluation}
We evaluate our model on simulated and real data sets. The fused distributions returned by the CFM are compared to those of the IFM and IOP and the base distributions. In addition, we compare the fusion performances to the performances of each classifier's meta classifier and the related method proposed by \citet{pirs2019}. 
As performance measures, we consider entropy and log-loss for quantifying uncertainty reduction through fusion and correctness of classifications.

\begin{figure*}[b]
	\begin{center}
		{\subfloat[SIM 1: $\bm{\alpha^1} {=} \bm{\alpha^2} {=} ((3,2,2),(2,3,2),(2,2,3))$ ]{\includegraphics[width=0.246\linewidth]{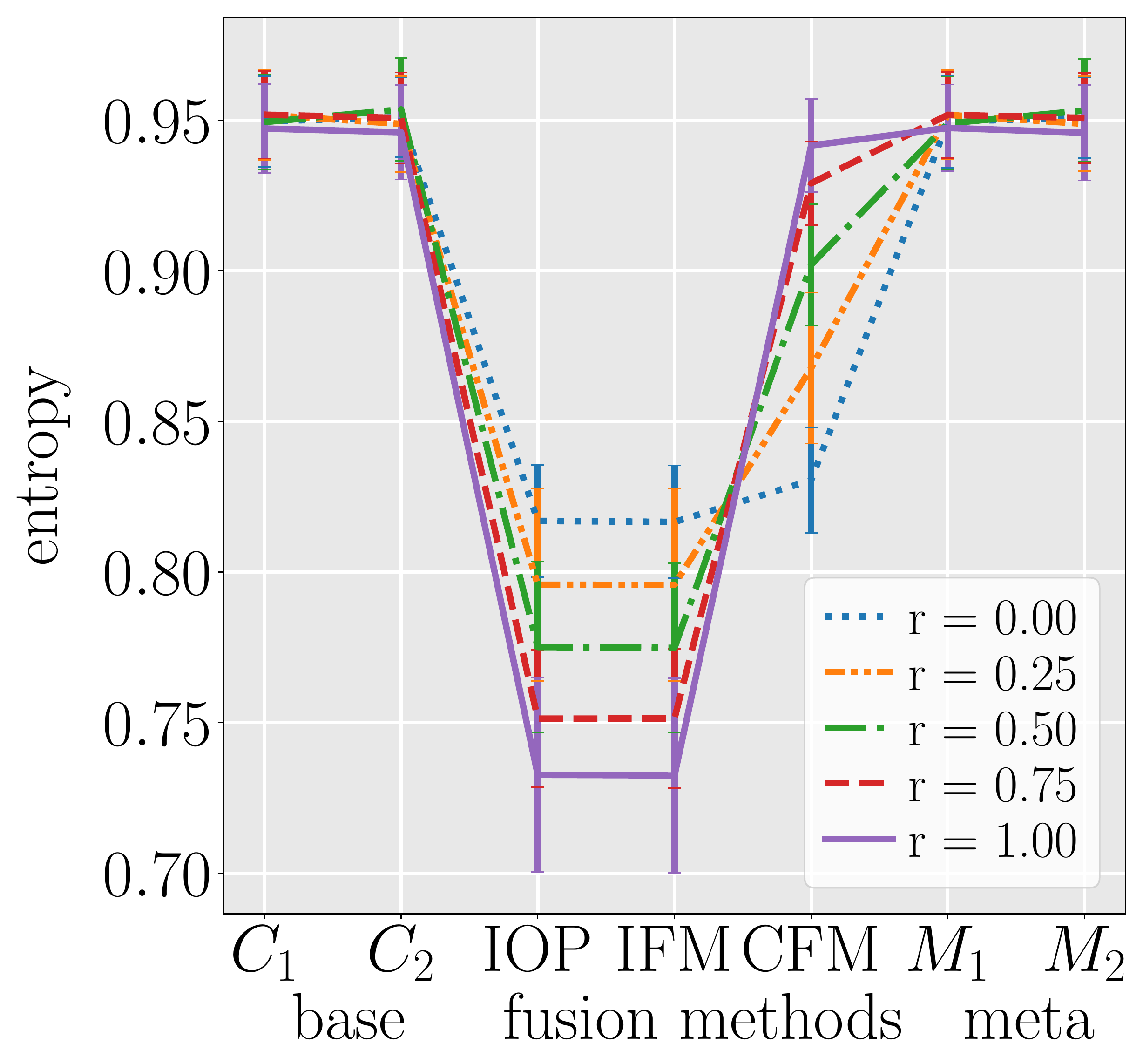}
				\includegraphics[width=0.24\linewidth]{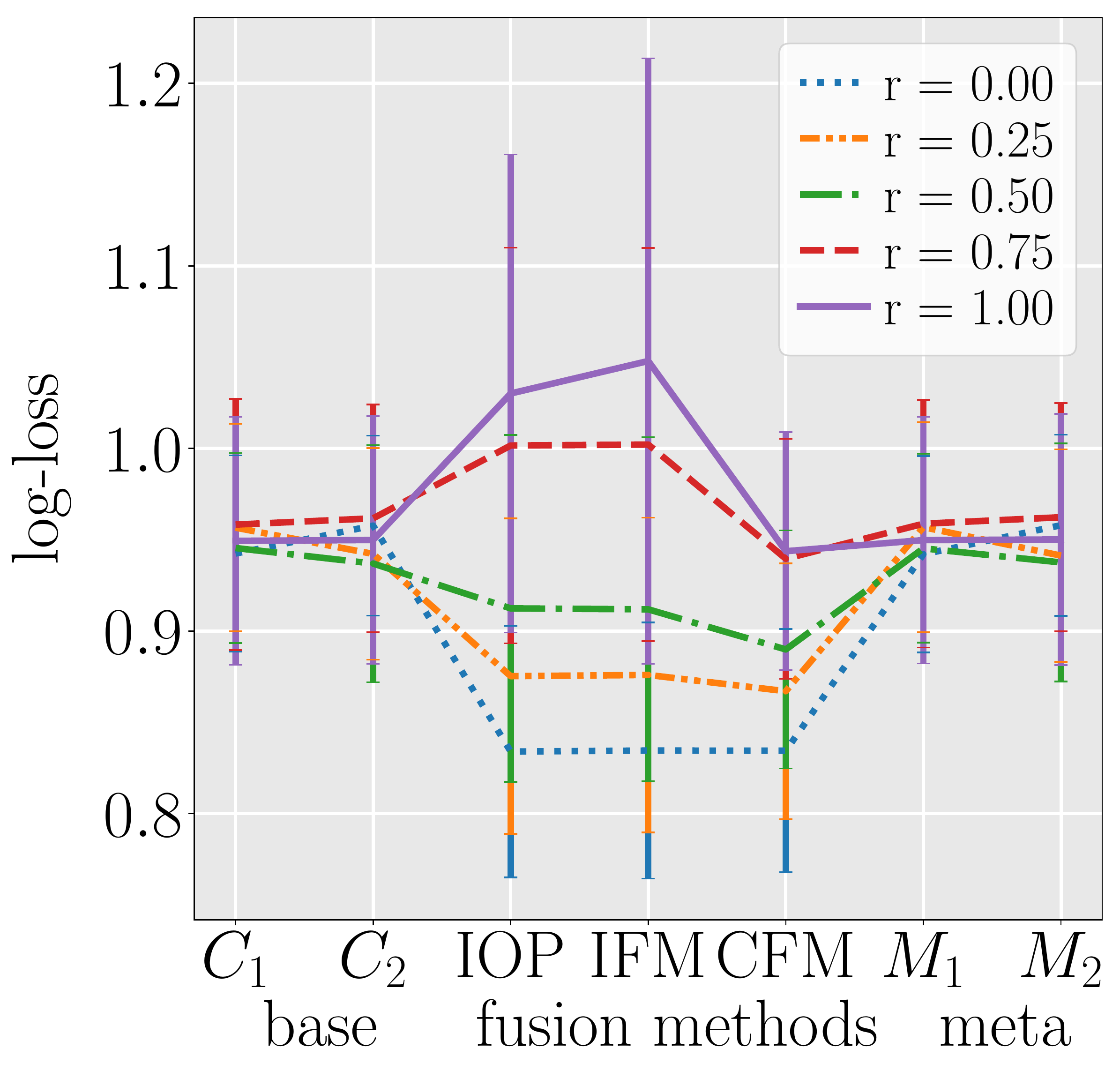}}}
		{\subfloat[SIM 2: $\bm{\alpha^1} {=} \bm{\alpha^2} {=} ((12,8,8),(8,12,8),(8,8,12))$]{\includegraphics[width=0.24\linewidth]{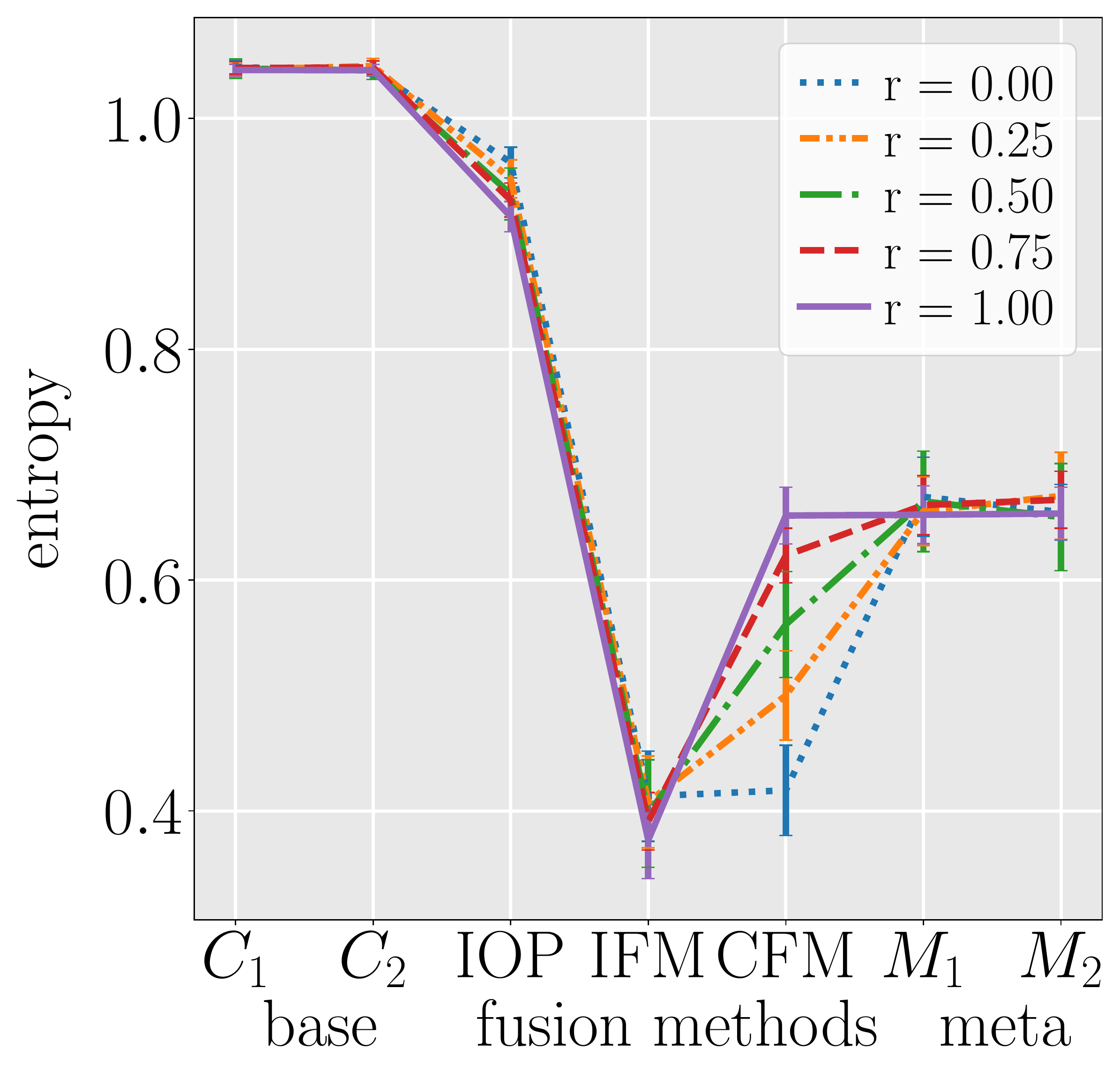}
				\includegraphics[width=0.24\linewidth]{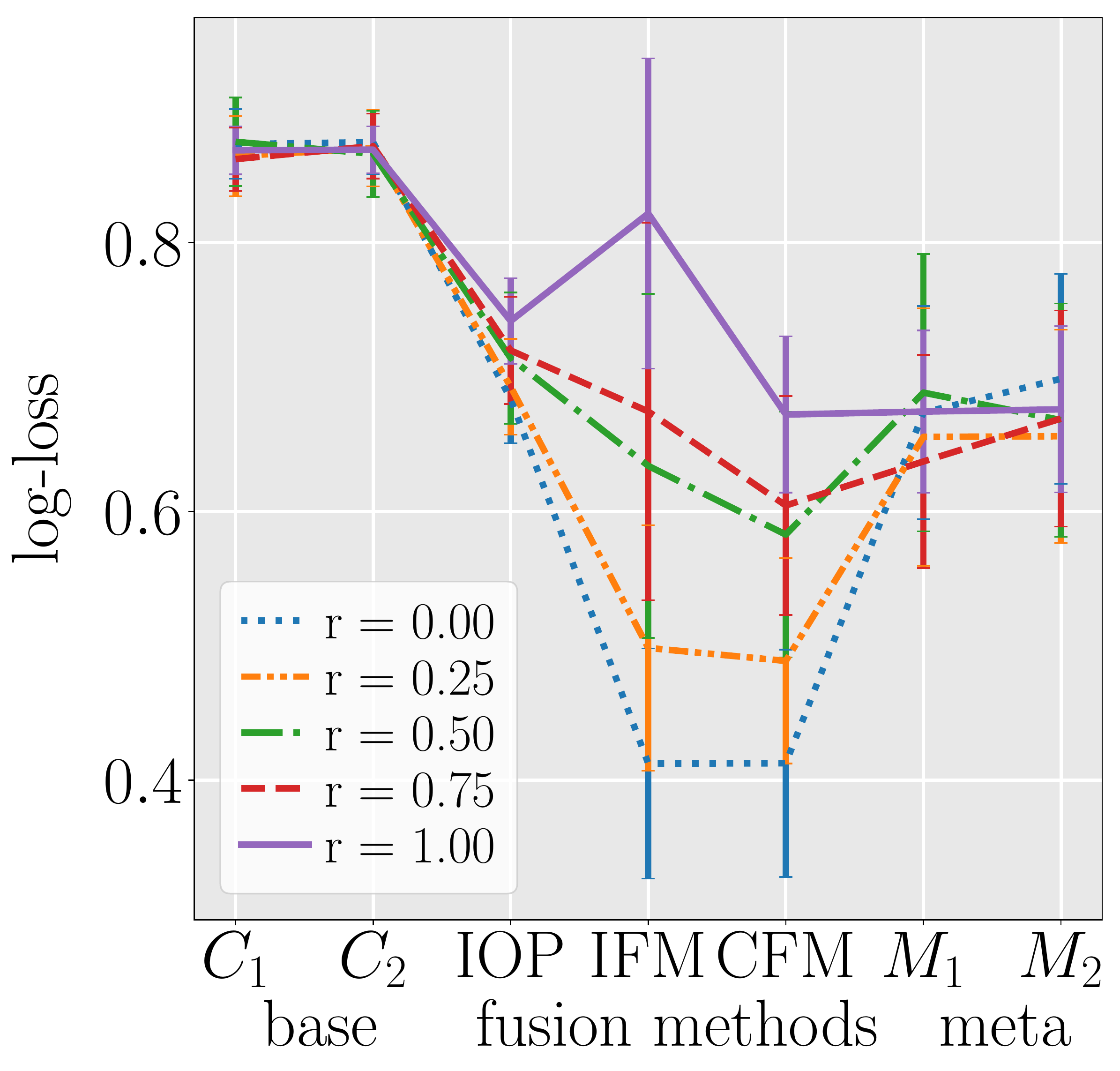}}}
		\caption{Fusion performances on simulated data in terms of mean entropy and log-loss. We compare the performance of base classifiers $C_1$, $C_2$, the three fusion methods IOP, IFM, and CFM, and the meta classifiers $M_1$, $M_2$. We show the fusion behavior for five levels of correlation between the base classifiers and different marginal model parameters, implying IOP fusion (a) and higher reinforcement due to decreased classifier variance and uncertainty (b). Standard deviations are shown as error bars.}
		\label{fig:results_sim}
	\end{center}
\end{figure*}

%%%%%%%%%%%%%%%%%%%%%%%%%%%%%%%%%%%%%%%%%%%%%%%%%%%%%%%%%%%%%%%%%%%%%%%%%%%%%%%%%%%%%
%%%%%%%%%%%%%%%%%%%%%%%%%%%%%%%%%%%%%%%%%%%%%%%%%%%%%%%%%%%%%%%%%%%%%%%%%%%%%%%%%%%%%
\subsection{Simulated data sets} \label{sec:eval_sim}
We created different simulated data sets by generating random samples of output distributions $\bm{x_i^1}$ and $\bm{x_i^2}$ of $K=2$ classifiers for different given marginal parameters $\bm{\alpha}$, correlation parameters $\bm{\delta}$ and true class labels $t_i$ with $J=3$ possible outcomes according to the generative model of the CFM (Figure \ref{fig:fusion_models}(b)).
To show the normative fusion behavior depending on the base classifiers' correlation, for two sets of marginal parameters $\bm{\alpha}$, we chose different correlation parameters $\bm{\delta}$ respectively that correspond to the correlations 0.0, 0.25, 0.5, 0.75, 1.0 between the two classifiers' outputs. For all five correlation levels, we generated 25 simulated random test sets on which we evaluate, each consisting of 60 test examples (20 per class) composed of two categorical distributions and their corresponding class label. Since the true parameters of the data were known, no training data were required.
We chose the marginal parameters to represent two prototype cases of classifier models in order to demonstrate that the effect of correlation on the fusion behavior also depends on the individual classifiers' marginal Dirichlet models. One of the chosen classifier models leads to IOP fusion for zero correlation, one represents two classifiers with decreased variance and uncertainty.

For the first simulated data set SIM 1, we determine the marginal parameters $\bm{\alpha}$ of the CFM such that it reduces to IOP if $r=0$. As shown in Figure \ref{fig:results_sim}(a), therefore, the results of IOP and the IFM are equal regarding entropy and log-loss.
The shown entropies reveal that the higher the correlation between the classifiers is, the more uncertainty is reduced by fusing with IOP or the IFM.
In contrast, when fusing with the CFM, we see less uncertainty reduction through fusion for higher correlations. Particularly, for $r=1$, there is no uncertainty reduction. The mean entropy is the same as for the two meta classifiers.
Also, the CFM's mean log-loss is equal to the meta classifiers' log-loss if $r=1$. Thus, as expected, we see no change in performance through fusion for highly correlated classifiers when using the CFM. Since we chose the marginals according to IOP fusion, the CFM's performance also equals the performances of the base classifiers.
In general, the CFM performs best at all correlation levels. Particularly for high correlations, it outperforms the other fusion methods, which assume independence, overestimate uncertainty reduction, and therefore perform even worse than the base classifiers.

The second simulated data set SIM 2 was generated setting the CFM's marginal parameters $\bm{\alpha}$ according to the example in (\ref{eq:nmeq}) with $n=m=4$, which leads to increased uncertainty reduction through fusion in comparison to IOP for independent classifiers, since the modeled base classifiers' variance and uncertainty is decreased.
Accordingly, Figure \ref{fig:results_sim}(b) shows significantly lower mean entropies for the IFM than for IOP for all correlation levels. In contrast, for the CFM, the fused distributions' mean entropy increases with the correlation such as for SIM 1. If $r=1$, the CFM again shows the same entropy as the two meta classifiers. Hence, the fusion of two highly correlated base classifiers does not reduce the uncertainty.
This is confirmed by the log-loss (Figure \ref{fig:results_sim}(b)). However, in contrast to SIM 1, here, the meta classifiers' performances are increased compared to the base classifiers, and uncertainty is reduced. Therefore, the CFM outperforms the base classifiers also for a correlation of $r=1$. Note that, again, the CFM achieves the lowest log-loss and thus the best performance for all correlation levels.

%%%%%%%%%%%%%%%%%%%%%%%%%%%%%%%%%%%%%%%%%%%%%%%%%%%%%%%%%%%%%%%%%%%%%%%%%%%%%%%%%%%%%
%%%%%%%%%%%%%%%%%%%%%%%%%%%%%%%%%%%%%%%%%%%%%%%%%%%%%%%%%%%%%%%%%%%%%%%%%%%%%%%%%%%%%
\subsection{Real data sets} \label{sec:eval_real}
In addition to simulated data sets, we also evaluated the CFM on 5 real data sets, Bookies A, Bookies B, DNA A, DNA B, DNA C.
Both Bookies data sets are composed of $K=2$ bookmakers' odds for football matches of the English Premier League\footnote{https://www.football-data.co.uk/englandm.php} (Bookies A) and the German Bundesliga\footnote{https://www.football-data.co.uk/germanym.php} (Bookies B). The target variable has $J=3$ possible outcomes, and for each match example, the odds were transformed to a 3-dimensional categorical probability distribution by normalizing their reciprocals. Thus, each bookie is considered as a base classifier and each example in the Bookies data sets is composed of two categorical distributions and a true class label. The correlation between the bookmakers' predictions is approximately 1 in both data sets; it ranges from 0.955 to 0.996 in different dimensions and for different values of $t_i$.

The DNA data set from the StatLog project\footnote{https://archive.ics.uci.edu/ml/datasets/Molecular+
	Biology+(Splice-junction+Gene+Sequences)} with a target variable with $J=3$ possible outcomes was used to construct three more data sets for evaluating the CFM. For each, we trained $K=2$ different classifiers on this data set. Their categorical output distributions on the corresponding test data set form the respective data set DNA A, DNA B, DNA C.
For DNA A, we trained two highly correlated classifiers by using the same classification method (kNN) and same training data but different hyperparameters ($k=120$ and $k=150$). The correlation between both classifiers is approximately 1; it ranges from 0.962 to 0.986 for different dimensions and values of $t_i$.
For DNA B, we trained two classifiers using the same classification method (kNN, $k=50$) but different training data. Their correlation ranges from 0.463 to 0.709 for different dimensions and values for $t_i$.
DNA C was created by training two different classifiers, a kNN classifier
($k=50$) and a Random Forest classifier, on the same training set. The
correlation between their output distributions ranges from 0.5 to 0.693 in different dimensions and for different values of $t_i$.

\begin{figure}
	\CenterFloatBoxes
	\begin{floatrow}
		\ffigbox
		{\includegraphics[width=0.49\linewidth]{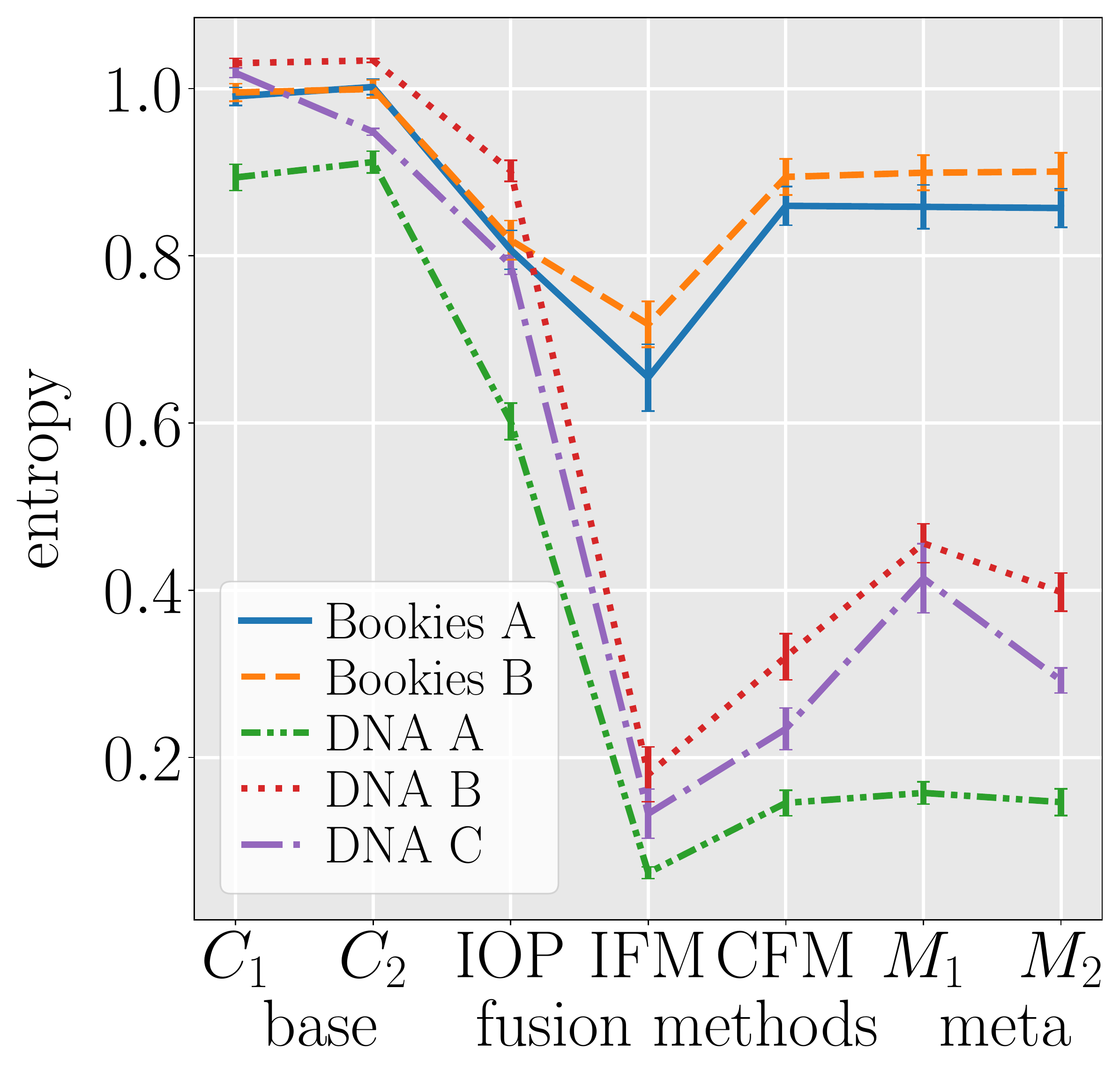}
			\includegraphics[width=0.49\linewidth]{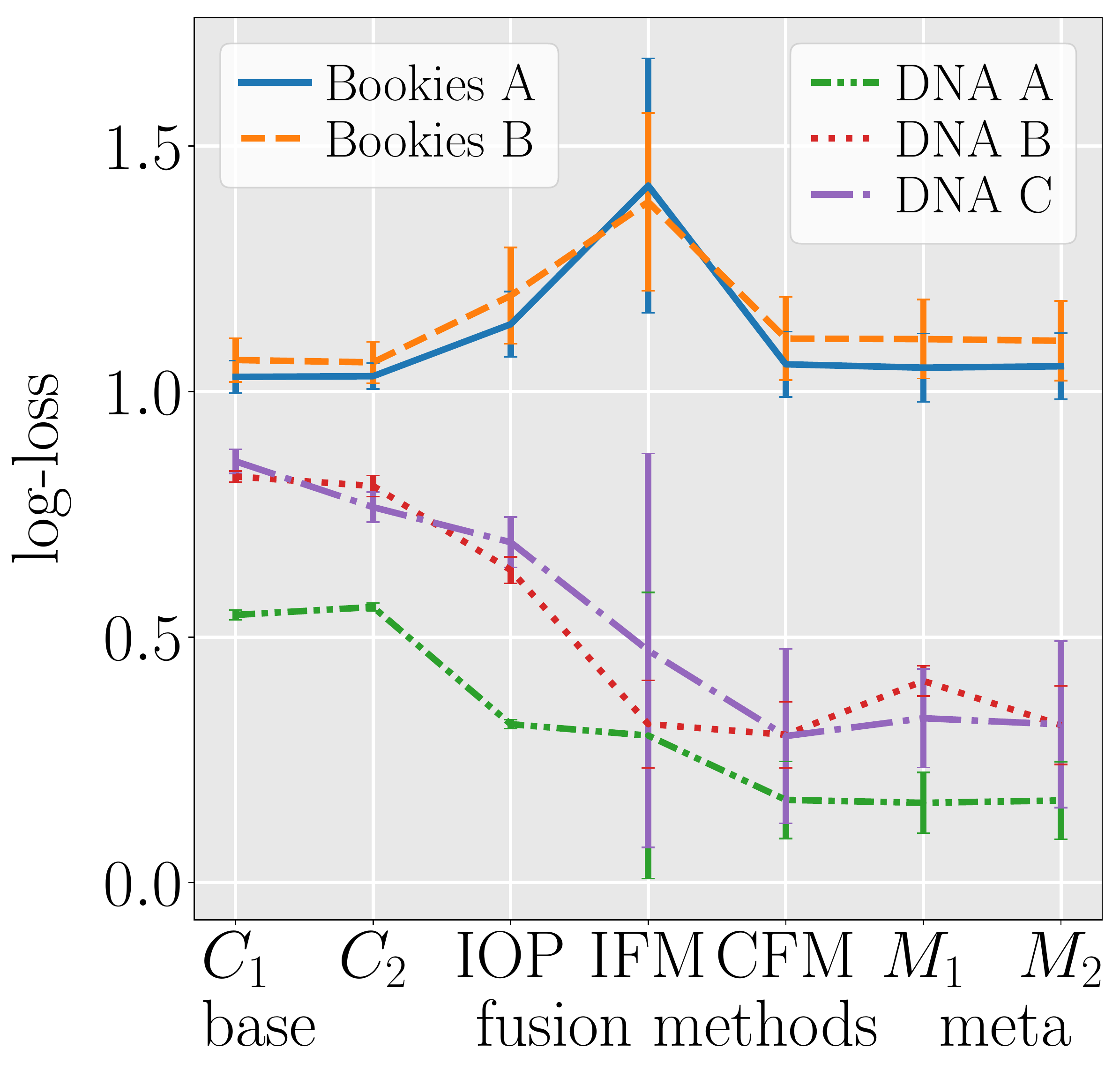}}
		{\caption{Fusion performances on real data in terms of mean entropy and log-loss. We compare the performance of base classifiers $C_1$, $C_2$, the fusion methods IOP, IFM, and CFM, and the meta classifiers $M_1$, $M_2$. Standard deviations are shown as error bars.}
		\label{fig:results_real}}
		\ttabbox
		{\begin{tabular}{l c c c c}
				\hline
				data set & CFM ($\mu\pm \sigma$) & Pirs ($\mu\pm \sigma$) \\
				\hline
				SIM 1 \tiny r=0.0 & $0.834\pm0.067$  & $0.915\pm0.03$ \\
				SIM 1 \tiny r=0.5 & $0.89\pm0.065$ & $0.938\pm0.039$ \\
				SIM 1 \tiny r=1.0 & $0.944\pm0.065$ & $0.96\pm0.056$ \\
				SIM 2 \tiny r=0.0 & $0.412\pm0.085$ & $0.582\pm0.048$ \\
				SIM 2 \tiny r=0.5 & $0.583\pm0.092$ & $0.66\pm0.065$ \\
				SIM 2 \tiny r=1.0 & $0.672\pm0.058$ & $0.717\pm0.041$ \\
				Bookies A & $1.056\pm0.067$ & $1.165\pm0.035$ \\
				Bookies B & $1.108\pm0.085$ & $1.176\pm0.052$ \\
				DNA A & $0.169\pm0.078$ & $0.177\pm0.021$ \\
				DNA B & $0.301\pm0.067$ & $0.421\pm0.043$ \\
				DNA C & $0.298\pm0.178$ & $0.351\pm0.092$ \\
				Bookies C & $1.056\pm0.056$ & $1.297\pm0.046$ \\
				\hline
			\end{tabular}
		}
		{\caption{Comparison of log-losses of the CFM and Pirs' method \citep{pirs2019} on simulated and real data. }
		\label{tab:pirs_comparison}}
	\end{floatrow}
\end{figure}

We randomly split each real data set into test and training set, while the test set contains 60 examples (20 per class) and the training set all remaining ones. On each random training split the model parameters $\bm{\alpha}$ and $\bm{\delta}$ were inferred, which were then used to fuse the distributions in the test set. The random splitting was repeated five times with different random seeds, and expectations and standard deviations of the resulting performance measures were computed, which are shown in Figure \ref{fig:results_real}.

For the three highly correlated data sets, Bookies A, B, DNA A, the CFM's performance is equal to the performances of the meta classifiers, both regarding entropy and log-loss. Thus, also on real data we confirm that fusion causes no uncertainty reduction and no change in performance if the base classifiers are highly correlated. However, this does not necessarily result in equal performances of the CFM and the base classifiers. Depending on the Dirichlet models learned for the individual classifiers, the CFM can still outperform highly correlated base classifiers, which we see for DNA A. Also, the CFM can perform worse than the base classifiers, e.g. for Bookies B, which is an effect of too similar Dirichlet models for different class labels $t_i$, as also noticed by \citet{pirs2019}.

For the less correlated data sets, DNA B and C, we see that the CFM reduces less uncertainty than the IFM but is more certain than the meta classifiers. Also, the CFM performs best of all fusion methods and better than base and meta classifiers.

%%%%%%%%%%%%%%%%%%%%%%%%%%%%%%%%%%%%%%%%%%%%%%%%%%%%%%%%%%%%%%%%%%%%%%%%%%%%%%%%%%%%%
%%%%%%%%%%%%%%%%%%%%%%%%%%%%%%%%%%%%%%%%%%%%%%%%%%%%%%%%%%%%%%%%%%%%%%%%%%%%%%%%%%%%%
\subsection{Comparison to the approach by Pirs and Strumbelj} \label{sec:pirs}
The model introduced by \citet{pirs2019}, which relies on modeling transformed classifier outputs with a multivariate normal mixture, is the only comparable Bayesian method for fusing correlated probabilistic classifiers.
Contrary to \citet{pirs2019}, on simulated and real data we show that although fusion should not reduce the uncertainty if $r=1$, in a normative framework fused classifiers can outperform highly correlated base classifiers due to the models learned for the individual classifiers. Moreover, we additionally compared the performances of the CFM and Pirs' model in terms of log-loss. As can be seen in Table \ref{tab:pirs_comparison}, the CFM outperforms on all tested data sets. The simulated data sets not displayed in the table showed similar results but are left out for brevity.
In addition to the real data sets discussed in Section \ref{sec:eval_real} we also compared the CFM and Pirs' model on an additional data set equivalent to Bookies A but with $K=3$ bookmakers, Bookies C. Also on this data set, the CFM outperform Pirs' method.

%%%%%%%%%%%%%%%%%%%%%%%%%%%%%%%%%%%%%%%%%%%%%%%%%%%%%%%%%%%%%%%%%%%%%%%%%%%%%%%%%%%%%
%%%%%%%%%%%%%%%%%%%%%%%%%%%%%%%%%%%%%%%%%%%%%%%%%%%%%%%%%%%%%%%%%%%%%%%%%%%%%%%%%%%%%
\section{Conclusion} \label{sec:conclusion}
In this work, we introduced a Bayesian model of classifier fusion based on a new correlated Dirichlet distribution. We derived Bayes optimal fusion behavior for probabilistic classifiers that output categorical distributions, which considers the classifiers' bias, variance, uncertainty, and correlation. We showed that uncertainty reduction through fusion should be the lower, the higher the correlation between the classifiers is, resulting in no uncertainty reduction through fusion if $r=1$. However, this does not necessarily lead to equal performances of the fused classifier and the base classifiers if a model for each classifier is learned. 
The proposed normative fusion model offers a new perspective on Bayesian combination of probabilistic classifiers,
thereby clarifying how the correlation between classifiers affects uncertainty reduction through fusion and subsuming well known pioneering expert opinion aggregation techniques.
Since it additionally outperforms the only comparable model it should be the method of choice if Bayes optimal fusion is the goal.
However, as classification could potentially be used in conjunction with data and tasks with negative societal impact, we encourage responsible deployment of the proposed approach.

\begin{ack}
This work has been funded by the German Federal Ministry of Education and
Research (BMBF) (projects Kobo34 [16SV7984] and IKIDA [01IS20045]).
Additionally, we acknowledge support by the Hessian Ministry of Higher Education, Research, Science, and the Arts (HMWK) (projects "The Third Wave of AI" and "The Adaptive Mind") and the Hessian research priority program LOEWE within the project "WhiteBox".

\end{ack}

\small
\bibliography{references}
\bibliographystyle{plainnat}
\normalsize

\end{document}